\documentclass[cameraready]{Interspeech}
\title{WavSLM: Single-Stream Speech Language Modeling via WavLM Distillation}

\author[affiliation={1,2}, orcid=0000-0001-6088-2410, correspondingauthor]{Luca}{Della Libera}
\author[affiliation={3,1,2}, orcid=0000-0002-7593-6589]{Cem}{Subakan}
\author[affiliation={1,2}, orcid=0000-0002-3929-5526]{Mirco}{Ravanelli}
\address{
    $^1$Concordia University \, \,
    $^2$Mila-Quebec AI Institute \, \,
    $^3$Université Laval
}

\email{luca.dellalibera@mail.concordia.ca, cem.subakan@ift.ulaval.ca, mirco.ravanelli@concordia.ca}


\usepackage{comment}
\usepackage{multirow, makecell}
\usepackage{pifont}
\usepackage{cite}
\usepackage[capitalize,noabbrev]{cleveref}
\usepackage{microtype}
\newcommand{\cmark}{\ding{51}}
\newcommand{\xmark}{\ding{55}}

\renewcommand{\footnote}[1]{}
\begin{document}

\maketitle

\begin{abstract}
Large language models show that simple autoregressive training can yield scalable and coherent generation, but extending this paradigm to speech remains challenging due to the entanglement of semantic and acoustic information. Most existing speech language models rely on text supervision, hierarchical token streams, or complex hybrid architectures, departing from the single-stream generative pretraining paradigm that has proven effective in text. In this work, we introduce WavSLM, a speech language model trained by quantizing and distilling self-supervised WavLM representations into a single codebook and optimizing an autoregressive next-chunk prediction objective. WavSLM jointly models semantic and acoustic information within a single token stream without text supervision or text pretraining. Despite its simplicity, it achieves competitive performance on consistency benchmarks and speech generation while using fewer parameters, less training data, and supporting streaming inference.\looseness=-1
\end{abstract}

\section{Introduction}
Large language models (LLMs) have transformed natural language processing by showing that scalable, coherent, and controllable generation can emerge from a simple next-token prediction objective~\cite{dubey2024llama3herdmodels,jiang2024mixtralexperts,comanici2025gemini25,singh2025openaigpt5card,deepseekai2025deepseekv3}. Extending this paradigm to speech, however, remains a fundamental challenge. Unlike text, speech is a high-dimensional, continuous signal that entangles semantic, prosodic, and acoustic information across multiple time scales. This intrinsic complexity makes both representation learning and long-range sequence modeling substantially more difficult than in text-based language modeling.

Speech language models (SLMs)~\cite{arora2025landscapespokenlanguagemodels} address this challenge by converting waveforms into sequences of discrete tokens using \textbf{neural audio codecs}~\cite{mousavi2025dates,guo2025discretespeechtokensreview} and modeling these sequences autoregressively. While conceptually similar to text tokenization, speech tokenization is inherently more complex, as it must simultaneously preserve semantic content and acoustic detail while remaining efficient, streamable, and low-bitrate.

Early work on audio language modeling primarily focused on semantic modeling, quantizing self-supervised speech representations~\cite{baevski2020wav2vec2,hsu2021hubert,chen2022wavlm} into discrete units~\cite{lakhotia2021gslm, kharitonov2022pgslm}. Although these models produced intelligible speech, they continued to lag behind text-based language models in terms of factual knowledge, long-range consistency, and controllability. To address these limitations, speech-text models incorporate text supervision or pretrained text language models, either by extending text LLMs with audio tokens~\cite{hassid2023textually,maiti2024voxtlm,park2024speechssm,cuervo2025latefusionmultilevelfission}, by interleaving text and speech representations~\cite{nguyen2024spiritlminterleavedspokenwritten}, or by training joint speech-text objectives~\cite{turetzky2024lastlanguagemodelaware,lu2025latentspeechtexttransformer}.

Beyond semantics, accurately modeling acoustic variation, such as speaker identity and fine-grained prosody, is essential for high-quality speech generation. Hierarchical generation frameworks address this limitation by explicitly modeling acoustic tokens conditioned on semantic representations, enabling improved speaker generalization and acoustic coherence~\cite{borsos2023audiolm, rubenstein2023audiopalmlargelanguagemodel}.
More recent work has explored unified semantic-acoustic modeling. Moshi~\cite{defossez2024moshi} bridges semantic and acoustic representations using dedicated codebooks and hybrid architectures that combine temporal and depth-wise transformers, albeit at the cost of significant architectural complexity. LLaMA-Mimi~\cite{sugiura2025llamamimispeechlanguagemodels}, similarly to \cite{nguyen2024spiritlminterleavedspokenwritten}, simplifies this design by interleaving semantic and acoustic tokens within a single sequence, although tokens at different time steps still serve distinct semantic or acoustic roles.

Despite recent progress, most semantic-acoustic SLMs rely on hybrid pipelines that incorporate text-pretrained LLMs, multiple token streams modeled partly autoregressively and partly in parallel, or additional text-alignment objectives such as direct text supervision or word boundaries. These design choices deviate from the single-stream generative pretraining paradigm that has proven effective in text for enabling strong generalization and emergent capabilities. While scaling model size and training data can partially compensate for this complexity~\cite{coreteam2025mimoaudioaudiolanguagemodels}, such gains typically require several orders of magnitude more data and compute~\cite{cuervo2024scaling}, raising concerns about efficiency and scalability. This naturally raises a fundamental question: \emph{can comparable performance instead be achieved through better representations, rather than increased scale and architectural complexity?}\looseness=-1

In this work, we investigate speech language modeling using a {single discrete codebook} that jointly represents semantic and acoustic information, without relying on text supervision or text-pretrained models. By closely mirroring the text paradigm of simple autoregressive training with a single decoder, our study isolates the role of the representation itself. We hypothesize that sufficiently expressive speech representations can support effective language modeling within a single-stream, single-decoder framework.
\begin{figure*}[t!]
  \centering
  \includegraphics[width=0.69\textwidth]{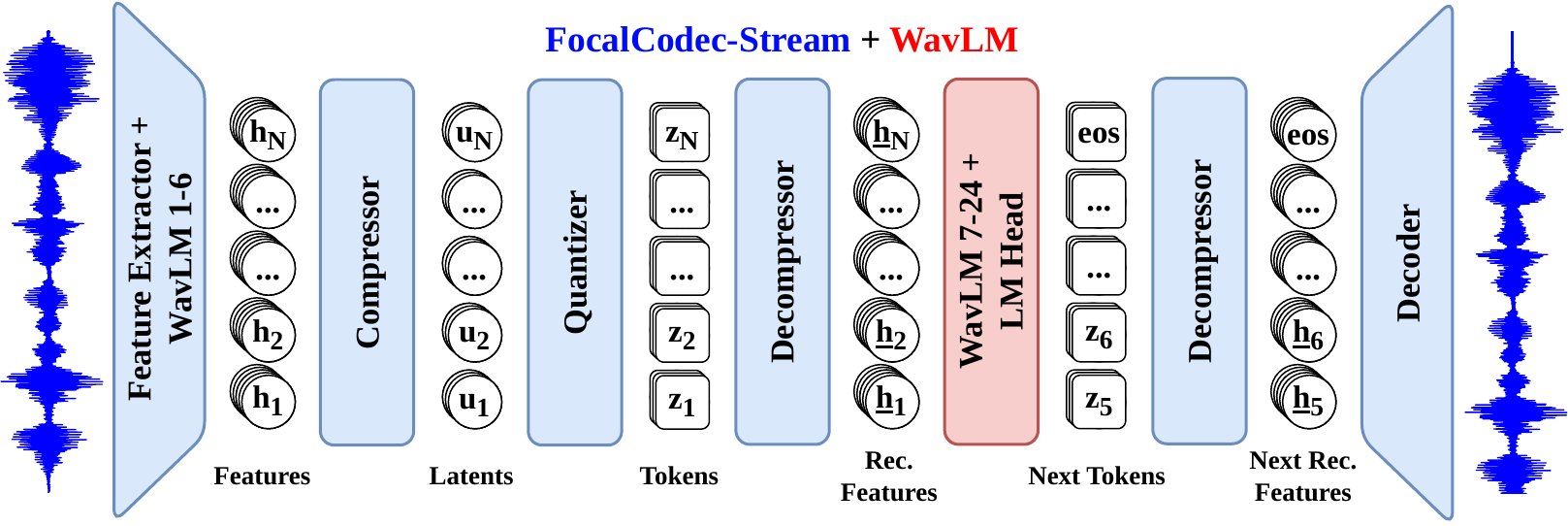}
  \vspace{-0.4cm}
  \caption{\footnotesize WavSLM architecture. Raw speech is processed by FocalCodec-Stream \textcolor{blue}{(blue)}, which includes the feature extractor and lower WavLM layers, followed by a compressor, quantizer, decompressor, and decoder to produce a low-bitrate, single-stream sequence of discrete tokens. The decompressor converts tokens back into continuous features that are compatible with the upper WavLM layers \textcolor{red}{(red)}. These layers are used as a causal speech language modeling backbone, with a lightweight language modeling head on top. The model is trained with a next-chunk prediction objective, jointly predicting $C=4$ tokens at each autoregressive step.}
  \label{fig:wavslm}
\vskip -0.6cm
\end{figure*}
In particular, our contributions are threefold:
\begin{itemize}
    \item We introduce \textbf{WavSLM}, an SLM trained by quantizing and distilling \textbf{WavLM representations} into an autoregressive next-chunk predictor over a single stream of tokens. To the best of our knowledge, this is the first SLM that jointly captures semantic and acoustic information using a \textbf{single codebook}, without hierarchical or multi-stream tokenization.
    
    \item We show that WavSLM achieves competitive performance on both semantic and acoustic consistency tasks, as well as on speech generation, while being significantly smaller and trained on substantially less data than large-scale SLMs. Unlike most related work, our model is trained \textbf{exclusively on speech}, without any text supervision or text-pretraining, and is fully streamable, enabling real-time speech generation.
    
    \item We analyze several design factors for single-stream speech language modeling, including context window size, chunk size, and vocabulary size, and study the trade-offs involved in jointly modeling semantic and acoustic information.
\end{itemize}
Code and checkpoints will be released publicly to foster reproducibility and further research. \textbf{Demo samples are available at \href{https://lucadellalib.github.io/wavslm-web/}{https://lucadellalib.github.io/wavslm-web/}}.

\section{WavSLM}
\label{sec:wavslm}

\subsection{Speech Representations and Tokenization}
A central design choice in the proposed {WavSLM} framework (see \cref{fig:wavslm}) is to build on {WavLM}~\cite{chen2022wavlm} representations. WavLM provides rich, hierarchical speech features spanning multiple levels of abstraction, ranging from low-level acoustic cues in early layers to increasingly semantic representations in deeper layers. Such a hierarchical structure is particularly well-suited for speech-language modeling, which must jointly capture lexical content, speaker identity, prosody, and other paralinguistic factors.
In particular, we use the representations from the \textbf{6-th transformer layer} of WavLM-large\footnote{\href{https://github.com/microsoft/unilm/tree/master/wavlm}{https://github.com/microsoft/unilm/tree/master/wavlm}}, which strike a balance between semantic richness and fine-grained acoustic detail~\cite{baas2023knnvc,mousavi2024how,dellalibera2025focalcodec,dellalibera2026dycast}. Our hypothesis is that these mid-level representations are sufficiently expressive to enable autoregressive modeling of speech, without relying on text as an intermediate representation.

Rather than learning a tokenizer for these representations from scratch, we leverage \textbf{FocalCodec-Stream}\footnote{\href{https://github.com/lucadellalib/focalcodec}{https://github.com/lucadellalib/focalcodec}}~\cite{dellalibera2025focalcodecstream}, a streamable neural speech codec that operates directly on WavLM-6 features.
FocalCodec-Stream builds on a causally distilled WavLM-6 encoder, replacing standard convolutions with causal convolutions and full-context gated relative attention with sliding-window gated relative chunked attention to enable streaming. The encoder extracts features that capture both acoustic and semantic information. These features are mapped to a low-dimensional space by a causal compressor based on focal modulation~\cite{yang2022focalnets,dellaliber2024focal}, followed by a single-codebook binary spherical quantizer~\cite{zhao2024bsq}, producing a \textbf{single stream of discrete tokens} at 50 Hz. A mirrored causal decompressor, which includes a chunk-wise feed-forward refiner module, reconstructs the sequence of continuous features, while a causal WaveNeXt~\cite{okamoto2023wavenext} decoder resynthesizes the waveform. The system operates with a chunk size of 4 tokens, yielding a theoretical streaming latency of 80 milliseconds.

Importantly, the decompressed tokens can be projected back into a continuous feature space that remains compatible with the upper layers of WavLM. Since FocalCodec-Stream is optimized for feature reconstruction, the decoded features can be viewed as approximations of the original WavLM representations. This design allows FocalCodec-Stream to act as an {interface} between raw speech and higher-level semantic representations: discrete tokens serve as the modeling units, while reconstructed features preserve access to the hierarchical WavLM feature space.

\begin{table*}[t]
\setlength{\tabcolsep}{4pt}
\caption{\footnotesize Likelihood-based evaluation. \textbf{Best} and \underline{second-best} results are highlighted. Human scores are {\setlength{\fboxsep}{1pt}\fbox{boxed}} when they outperform all models.}
\vspace{-19pt}
\label{tab:acoustic_semantic_results}
\begin{center}
\begin{footnotesize}
\resizebox{0.89\textwidth}{!}{%
\begin{tabular}{lccccccccccc}
\toprule
\multirow{2}{*}[-0.7ex]{\textbf{Model}}
& \multirow{2}{*}[-0.7ex]{\textbf{Params}}
& \multirow{2}{*}[-0.7ex]{\makecell{\textbf{Textually} \\ \textbf{Pretrained}}}
& \multirow{2}{*}[-0.7ex]{\textbf{Codebooks}}
& \multicolumn{3}{c}{\textbf{Acoustic Consistency}} 
& \textbf{Alignment}
& \multicolumn{3}{c}{\textbf{Spoken Content}}
& \multirow{2}{*}[-0.7ex]{\textbf{Avg} $\uparrow$} \\
\cmidrule(lr){5-7}
\cmidrule(lr){8-8}
\cmidrule(lr){9-11}
& & & &
Sent. $\uparrow$
& Spk. $\uparrow$
& Gend. $\uparrow$
& Sent. $\uparrow$
& sWUGGY $\uparrow$
& sBLiMP $\uparrow$
& tSC $\uparrow$
& \\
\midrule
Human~\cite{hassid2023textually, maimon2025salmon}
& -- & -- & --
& {\setlength{\fboxsep}{1pt}\fbox{97.2}} & {\setlength{\fboxsep}{1pt}\fbox{91.5}} & {\setlength{\fboxsep}{1pt}\fbox{98.6}} & {\setlength{\fboxsep}{1pt}\fbox{93.3}} & -- & -- & {\setlength{\fboxsep}{1pt}\fbox{90.2}} & -- \\

\midrule
\multirow{10}{*}{\rotatebox{90}{\scriptsize\textbf{Large-Scale Baselines}}}

\hspace{0.05cm} TWIST 1.3B~\cite{hassid2023textually, maimon2025salmon}
& 1.3B & \cmark & 1 $\times$ 500
& 61.5 & 69.0 & 69.5 & \textbf{53.0} & \underline{72.7} & 57.0 & 70.6 & 64.8 \\
\hspace{0.35cm} TWIST 7B~\cite{hassid2023textually, maimon2025salmon}
& 7.0B & \cmark & 1 $\times$ 500
& 61.5 & 71.0 & 70.0 & 51.5 & \textbf{74.5} & {59.2} & {76.4} & 66.3 \\

\hspace{0.35cm} SpiRit LM~\cite{nguyen2024spiritlminterleavedspokenwritten, maimon2025salmon}
& 7.0B & \cmark & 1 $\times$ 501
& 54.5 & 69.5 & 67.0 & 48.0 & 69.0 & 58.3 & {82.9} & 64.2 \\
\hspace{0.35cm} SpiRit LM Expr.~\cite{nguyen2024spiritlminterleavedspokenwritten, maimon2025salmon}
& 7.0B & \cmark & [501, 64, 100]
& 73.5 & 81.0 & 85.0 & \underline{52.0} & 65.0 & 54.2 & 75.4 & \underline{69.4} \\
\hspace{0.35cm} Moshi 7B~\cite{defossez2024moshi}
& 7.7B & \cmark & 8 $\times$ 2048
& -- & -- & -- & -- & 72.6 & 58.8 & \underline{83.0} & -- \\
\hspace{0.35cm} LAST 1.3B~\cite{turetzky2024lastlanguagemodelaware,maimon2025salmon} & 1.3B & \cmark & 1 $\times$ 500 & 65.0 &  64.5 & 68.5 & 53.5 & -- & -- & -- & -- \\
\hspace{0.35cm} SpeechSSM 2B~\cite{park2024speechssm}
& 2.0B & \cmark & 1 $\times$ 32000
& -- & -- & -- & -- & 55.8 & \underline{60.9} & -- & -- \\

\hspace{0.35cm} SmolTolk 2B~\cite{cuervo2025latefusionmultilevelfission}
& 2.0B & \cmark & 1 $\times$ 500
& -- & -- & -- & -- & -- & \textbf{61.9} & \textbf{87.6} & -- \\

\hspace{0.35cm} LLaMA-Mimi 1.3B~\cite{sugiura2025llamamimispeechlanguagemodels}
& 1.3B & \cmark & 4 $\times$ 2048
& \textbf{79.0} & 85.0 & 83.5 & 48.5 & 68.7 & 54.3 & 64.0 & 69.0 \\
\hspace{0.35cm} LLaMA-Mimi 8B~\cite{sugiura2025llamamimispeechlanguagemodels}
& 8.0B & \cmark & 4 $\times$ 2048
& 76.5 & 86.5 & 85.5 & 46.5 & 68.8 & 55.1 & 67.6 & \textbf{69.5} \\

\midrule
\multirow{8}{*}{\rotatebox{90}{\scriptsize\textbf{Data-Matched Baselines}}}

\hspace{0.1cm} HuBERT 25Hz~\cite{mousavi2025dates}
& $\sim$357M & \cmark & 1 $\times$ 500
& 62.5 & 69.0 & 69.5 & \textbf{53.0} & 70.5 & \underline{60.9} & 71.5 & 65.3 \\
\hspace{0.35cm} Enc-SMA-24~\cite{mousavi2025dates}
& $\sim$357M & \cmark & 8 $\times$ 1024
& 56.5 & 65.0 & 70.5 & 50.0 & 51.3 & 51.1 & 48.2 & 56.1 \\
\hspace{0.35cm} DAC-SMA-16~\cite{mousavi2025dates}
& $\sim$357M & \cmark & 8 $\times$ 1024
& 60.0 & 77.0 & 81.0 & 50.0 & 50.7 & 51.5 & 51.5 & 60.2 \\
\hspace{0.35cm} ST-S-16~\cite{mousavi2025dates}
& $\sim$357M & \cmark & 8 $\times$ 1024
& 58.0 & 65.0 & 66.5 & 49.5 & 56.9 & 51.1 & 55.7 & 57.5 \\
\hspace{0.35cm} Mimi-S-24~\cite{mousavi2025dates}
& $\sim$357M & \cmark & 8 $\times$ 2048
& 71.5 & 78.0 & 77.5 & \underline{52.0} & 62.2 & 52.3 & 54.3 & 64.0 \\
\hspace{0.35cm} DWavL-S-16~\cite{mousavi2025dates}
& $\sim$357M & \cmark & 6 $\times$ 1000
& 70.0 & 86.5 & \textbf{92.0} & 49.0 & 69.1 & 54.0 & 62.4 & 69.0 \\
\hspace{0.35cm} SQ-SMA-16~\cite{mousavi2025dates}
& $\sim$357M & \cmark & 4 $\times$ 19683
& 64.0 & 84.5 & 83.0 & 50.5 & 51.4 & 51.6 & 55.1 & 62.9 \\
\hspace{0.35cm} WT-SMA-24~\cite{mousavi2025dates}
& $\sim$357M & \cmark & 1 $\times$ 4096
& \underline{78.5} & 69.0 & 81.5 & 50.5 & 54.6 & 51.2 & 52.8 & 62.6 \\

\midrule
\multirow{3}{*}{\rotatebox{90}{\scriptsize\textbf{Proposed}}}

\hspace{0.05cm} \textbf{WavSLM-2k}
& 305M & \xmark & 1 $\times$ 2048
& 75.5 & 86.0 & 89.5 & 45.5 & 63.9 & 54.6 & 62.9 & 68.3 \\
\hspace{0.35cm} \textbf{WavSLM-4k}
& 307M & \xmark & 1 $\times$ 4096
& 75.0 & \textbf{88.5} & \underline{90.5} & 51.5 & 63.7 & 53.9 & 63.3 & \textbf{69.5} \\
\hspace{0.35cm} \textbf{WavSLM-65k}
& 370M & \xmark & 1 $\times$ 65536
& 75.0 & \underline{87.5} & 84.0 & 45.0 & 61.4 & 52.1 & 60.4 & 66.5 \\

\bottomrule
\end{tabular}
}
\end{footnotesize}
\end{center}
\vspace{-20pt}
\end{table*}

\subsection{Speech Language Modeling}
\label{subsec:slm}
Building on the tokenization pipeline described above, we repurpose the remaining WavLM layers (7--24) as an SLM. Specifically, we make these layers causal by applying a causal attention mask and fine-tune them using an autoregressive language modeling objective. A lightweight language modeling head is added on top to map the resulting features to a distribution over the speech token vocabulary. Through this process, the upper portion of the WavLM stack is distilled into an SLM that operates directly on reconstructed codec features and autoregressively generates speech continuations in the discrete token space. Crucially, the entire distillation process is \textbf{speech-only}, without any text supervision or text-pretrained initialization: both the underlying WavLM representations and the tokenization pipeline are learned exclusively from speech, and the SLM itself is initialized directly from the WavLM checkpoint. In contrast to many existing SLMs that initialize their weights from pretrained text LLMs, all linguistic structure in WavSLM emerges directly from speech data, closely mirroring the single-modality pretraining paradigm of text language modeling.

To better align with the temporal resolution of the tokenizer (which is causal over chunks of 4 tokens) and improve generation efficiency, we adopt a \textbf{next-chunk prediction} objective rather than predicting individual tokens one at a time. At each autoregressive step, the model predicts a chunk of $C=4$ consecutive tokens, matching the chunk size of the tokenizer.
In practice, we implement this using chunked causal attention with chunk size $C$, using full attention within chunks and causal masking across chunks. The input token sequence is left-shifted by $C$ positions to construct the targets, so that at each position the model predicts the token $C$ steps ahead, rather than the immediate next token. This allows the model to attend to all tokens within the current chunk while maintaining causality across chunks.

This formulation reduces the number of autoregressive steps required during generation, leading to faster inference while preserving high-resolution tokenization at the input. Additionally, it decouples temporal resolution from modeling efficiency: while the tokenizer produces fine-grained tokens, the effective downsampling factor can be controlled at the language modeling level via the chunk size, rather than being fixed by the tokenizer architecture.

To enable real-time and unbounded speech generation, we employ a sliding-window attention mechanism, restricting each prediction step to attend only to a fixed-length context of past tokens. This bounds memory usage and computational complexity, allowing speech to be generated continuously with constant latency.

\section{Experimental Setup}
\label{sec:setup}
We train WavSLM on Libri-Light~\cite{kahn2020libri} ($\sim$60k hours of unlabeled speech). We train three variants of the model, corresponding to the three vocabulary sizes of FocalCodec-Stream (2k, 4k, and 65k).
During training, we randomly sample 30-second audio segments from each utterance. Validation is performed on LibriSpeech~\cite{panayotov2015librispeech} \texttt{dev-clean}.

WavSLM is initialized from the pretrained WavLM-large checkpoint for layers 7--24, while the linear language modeling head is randomly initialized. The model is trained using the next-chunk prediction objective described in \cref{subsec:slm}.
We do not use explicit begin-of-sequence (BOS) or end-of-sequence (EOS) tokens. Instead, training utterances are padded with zeros at the waveform level, which naturally correspond to silence in the token space. This design choice is compatible with continuous and streaming generation, where the language model can continue producing tokens corresponding to silence without requiring a hard termination signal.

Training is performed in mixed precision with a batch size of 16, using the AdamW~\cite{loshchilov2019adamw} optimizer with a learning rate of 0.0001, weight decay of 0.01, and gradient clipping with a maximum L2 norm of 5.0. The learning rate is reduced by a factor of 0.9 when the validation loss fails to improve by at least 0.0025. Training is stopped once no further improvement is observed for several consecutive epochs. In some runs, we found it beneficial to reset the learning rate after apparent convergence, which led to additional improvements. All models are trained on a single NVIDIA H100 GPU (80 GB).

\section{Results}

\subsection{Likelihood-Based Evaluation}
\begin{itemize}[label={},leftmargin=0pt]

\item \textbf{Tasks}.
We evaluate the acoustic and semantic modeling capabilities of WavSLM using likelihood-based benchmarks, following prior work on speech-only evaluation~\cite{arora2025landscapespokenlanguagemodels,mousavi2025dates,sugiura2025llamamimispeechlanguagemodels}. To assess acoustic and paralinguistic modeling, we adopt the SALMon benchmark~\cite{maimon2025salmon}, reporting results on sentiment, speaker, and gender consistency, as well as sentiment alignment. To evaluate semantic and linguistic knowledge, we use the ZeroSpeech~\cite{dunbar2021zero} variants of sWUGGY and sBLiMP, and further assess discourse-level coherence on Topic Story-Cloze (tSC)~\cite{mostafazadeh2016corpus,hassid2023textually}.

\begin{table}[t]
\setlength{\tabcolsep}{1.5pt}
\caption{\footnotesize Generation-based evaluation. \textbf{Best} and \underline{second-best} results are highlighted.}
\vspace{-9pt}
\label{tab:utmos_sim_ppl}
\centering
\begin{footnotesize}
\resizebox{0.995\linewidth}{!}{%
\begin{tabular}{lccccccc}
\toprule
\textbf{Model}
& \textbf{Params}
& \makecell{\textbf{Textually} \\ \textbf{Pretrained}}
& \textbf{Codebooks}
& \textbf{UTMOS} $\uparrow$
& \textbf{Sim} $\uparrow$
& \textbf{PPL} $\downarrow$
& \textbf{RTF} $\uparrow$ \\
\midrule
LLaMA-Mimi 1.3B~\cite{sugiura2025llamamimispeechlanguagemodels}
& 1.3B
& \cmark
& 4 $\times$ 2048
& 3.57
& 91.3
& \underline{153}
& 2.0 \\
LLaMA-Mimi 8B~\cite{sugiura2025llamamimispeechlanguagemodels}
& 8.0B
& \cmark
& 4 $\times$ 2048
& 3.56
& 91.5
& \textbf{122}
& 1.1 \\
\midrule
\textbf{WavSLM-2k}
& 305M
& \xmark
& 1 $\times$ 2048
& \textbf{3.72}
& \textbf{91.8}
& 161
& \textbf{5.9} \\
\textbf{WavSLM-4k}
& 307M
& \xmark
& 1 $\times$ 4096
& \underline{3.69}
& \underline{91.6}
& 162
& \underline{5.8} \\
\textbf{WavSLM-65k}
& 370M
& \xmark
& 1 $\times$ 65536
& 3.66
& 91.3
& 210
& \underline{5.8} \\
\bottomrule
\end{tabular}
}
\end{footnotesize}
%\vspace{-13pt}
\vspace{-18pt}
\end{table}

\begin{table*}[t]
\setlength{\tabcolsep}{2pt}
\caption{\footnotesize Effect of window and chunk size for WavSLM-4k. 
\textbf{Best} and \underline{second-best} results are highlighted.}
\vspace{-13pt}
\label{tab:likelihood_generative_metrics}
\begin{center}
\begin{footnotesize}
\resizebox{0.79\textwidth}{!}{%
\begin{tabular}{cccccccccccccc}
\toprule
\multirow{2}{*}[-0.7ex]{\makecell{\textbf{Window} \\ \textbf{Size}}}
& \multirow{2}{*}[-0.7ex]{\makecell{\textbf{Chunk} \\ \textbf{Size}}}
& \multicolumn{3}{c}{\textbf{Acoustic Consistency}} 
& \textbf{Alignment}
& \multicolumn{3}{c}{\textbf{Spoken Content}}
& \multirow{2}{*}[-0.7ex]{\textbf{Avg} $\uparrow$}
& \multicolumn{4}{c}{\textbf{Generation}} \\
\cmidrule(lr){3-5}
\cmidrule(lr){6-6}
\cmidrule(lr){7-9}
\cmidrule(lr){11-14}
& 
& Sent. $\uparrow$
& Spk. $\uparrow$
& Gend. $\uparrow$
& Sent. $\uparrow$
& sWUGGY $\uparrow$
& sBLiMP $\uparrow$
& tSC $\uparrow$
&
& UTMOS $\uparrow$
& Sim $\uparrow$
& PPL $\downarrow$
& RTF $\uparrow$ \\
\midrule

512 & 4
& 75.0 & \textbf{88.5} & \textbf{90.5} & \textbf{51.5}
& 63.7 & 53.9 & 63.3
& \textbf{69.5}
& \underline{3.69} & \underline{91.6} & 162 & 5.8 \\

1024 & 4
& \underline{76.0} & \underline{87.5} & \underline{90.0} & 47.0
& \textbf{64.8} & \underline{54.7} & \underline{66.7}
& \textbf{69.5}
& \underline{3.69} & \textbf{91.7} & \underline{151} & 5.8 \\

2048 & 4
& 74.5 & \textbf{88.5} & 88.5 & \underline{49.0}
& \underline{64.6} & \underline{54.7} & \textbf{66.9}
& \textbf{69.5}
& \textbf{3.70} & \textbf{91.7} & \textbf{148} & 5.8 \\

512 & 8
& \textbf{78.5} & 86.0 & 83.0 & 48.5
& 63.7 & \textbf{55.5} & 64.8
& \underline{68.6}
& 2.92 & 90.0 & 174 & \underline{10.9} \\

512 & 16
& 71.5 & 84.5 & 81.5 & \underline{49.0}
& 57.5 & \underline{54.7} & 62.7
& 65.9
& 1.97 & 86.5 & 181 & \textbf{16.4} \\

\bottomrule
\end{tabular}
}
\end{footnotesize}
\end{center}
\vspace{-20pt}
\end{table*}

\item \textbf{Baselines}.
We compare WavSLM against a diverse set of representative baselines. We include several large-scale SLMs, namely TWIST~\cite{hassid2023textually}, SpiRit LM~\cite{nguyen2024spiritlminterleavedspokenwritten}, Moshi~\cite{defossez2024moshi}, LAST~\cite{turetzky2024lastlanguagemodelaware}, SpeechSSM~\cite{park2024speechssm}, SmolTolk~\cite{cuervo2025latefusionmultilevelfission}, and LLaMA-Mimi~\cite{sugiura2025llamamimispeechlanguagemodels}. These baselines are substantially larger than our model variants (1.3--8B parameters versus 305--370M for WavSLM), are trained on several hundred thousand to millions of hours of speech, and rely on text pretraining by bootstrapping from large text LLMs. In contrast, WavSLM itself and FocalCodec-Stream are trained on $\sim$60k hours of speech. When also accounting for the data used to pretrain WavLM, the total amount of speech amounts to $\sim$94k hours. Importantly, WavSLM does not leverage any text pretraining.
To provide a more controlled comparison, we additionally include data-matched baselines from \cite{mousavi2025dates}, which are comparable to WavSLM in model size and trained on the same data. These models, however, still employ text pretraining, as they were initialized from Qwen-2.5 0.5B~\cite{qwen2025qwen25technicalreport}.

We report results from the original papers or subsequent works evaluating these models, leaving entries blank when unavailable. As all tasks are binary-choice accuracy benchmarks, we additionally report averaged scores where applicable. To choose between the two candidate continuations, we compute their WavSLM negative log-likelihood and select the one with the lower score.

\item \textbf{Discussion}.
As shown in \cref{tab:acoustic_semantic_results}, despite being substantially smaller and trained exclusively on speech, WavSLM performs competitively with large-scale, text-pretrained SLMs across most metrics. In particular, WavSLM-4k achieves the best average score among all proposed models and matches or surpasses several billion-parameter baselines, including LLaMA-Mimi and SpiRit LM variants, while using an order of magnitude fewer parameters and training data.

Across acoustic consistency tasks, WavSLM shows strong sensitivity to speaker identity, gender, and sentiment changes, with WavSLM-4k achieving top or second-best performance among data-matched and large-scale baselines. This suggests that a single codebook can preserve fine-grained acoustic attributes when built on expressive self-supervised representations. On acoustic-semantic alignment, WavSLM remains competitive, indicating that sentiment-related acoustic cues are consistently aligned with spoken content despite the absence of text supervision.

For spoken content evaluation, WavSLM achieves solid performance on sWUGGY, sBLiMP, and tSC, approaching or matching text-pretrained models trained at much larger scale. Notably, WavSLM outperforms all data-matched baselines on average, highlighting the effectiveness of distilling WavLM representations into an autoregressive next-chunk predictor. Overall, these results indicate that strong semantic and acoustic modeling can emerge from a single-stream, speech-only language model.
\end{itemize}

\subsection{Generation-Based Evaluation}
\begin{itemize}[label={},leftmargin=0pt]

\item \textbf{Tasks}.
To evaluate speech generation capabilities, we consider a continuation task where an audio prompt is provided and the model generates the remaining speech. We use utterances longer than 4 seconds from LibriSpeech~\cite{panayotov2015librispeech} \texttt{test-clean}, take the first half of the audio as a prompt, and generate a continuation of equal duration.
Following \cite{sugiura2025llamamimispeechlanguagemodels}, generation is performed using top-k sampling with a temperature of 0.8 and $k = 30$. To reduce variance, we generate 5 continuations per prompt and average all metrics across continuations and utterances in the test set.

We assess generation quality using three complementary metrics. First, we compute UTMOS~\cite{saeki2022utmos} to measure perceived naturalness. Second, we evaluate speaker consistency by computing the cosine similarity between speaker embeddings extracted from the prompt and the continuation using WavLM-base-SV\footnote{\scriptsize \href{https://huggingface.co/microsoft/wavlm-base-sv}{https://huggingface.co/microsoft/wavlm-base-sv}}~\cite{chen2022wavlm}. Third, we assess spoken content quality and coherence by computing GPT-2-large\footnote{\scriptsize \href{https://huggingface.co/openai-community/gpt2-large}{https://huggingface.co/openai-community/gpt2-large}}~\cite{radford2019gpt2} perplexity on transcriptions obtained with Whisper-large-v3\footnote{\scriptsize \href{https://huggingface.co/openai/whisper-large-v3}{https://huggingface.co/openai/whisper-large-v3}}~\cite{radford2022robust} after concatenating the prompt and the generated continuation. We additionally report the real-time factor, measured on an NVIDIA H100 GPU (80 GB).

\item \textbf{Baselines}.
We compare WavSLM against LLaMA-Mimi 8B, the strongest overall baseline in \cref{tab:acoustic_semantic_results} for which public checkpoints and inference code are available\footnote{\scriptsize \href{https://github.com/llm-jp/llama-mimi}{https://github.com/llm-jp/llama-mimi}}. Since checkpoints for the data-matched baselines are not publicly available, we also include LLaMA-Mimi 1.3B to provide a more size-comparable reference against our models.

\item \textbf{Discussion}.
The results are reported in \cref{tab:utmos_sim_ppl}. Despite being substantially smaller and trained without text pretraining, WavSLM achieves competitive generation quality across all metrics. In particular, WavSLM-2k obtains the best UTMOS and speaker similarity scores, indicating that the generated speech remains natural and preserves speaker characteristics from the prompt. The 4k variant shows similarly strong performance and provides a good overall balance across metrics.

In terms of spoken content quality, LLaMA-Mimi achieves lower perplexity, suggesting stronger linguistic modeling. Nevertheless, the gap remains moderate given the substantial differences in model size and training data scale between the models. Finally, WavSLM achieves substantially higher real-time factors than LLaMA-Mimi, reflecting significantly faster generation. This speed advantage stems from the smaller model size and the next-chunk generation strategy.

Interestingly, the 65k variant performs noticeably worse than the smaller-vocabulary models. We hypothesize that the larger vocabulary increases modeling complexity and requires more training data to be effectively learned under the current setup. We also note that, for fairness, we used the same generation hyperparameters from \cite{sugiura2025llamamimispeechlanguagemodels} and did not perform task-specific tuning for WavSLM, suggesting that further improvements may be possible with optimized decoding settings.
\end{itemize}

\subsection{Effect of Window and Chunk Size}
We analyze the effect of context window size and chunk size on WavSLM-4k in \cref{tab:likelihood_generative_metrics}. 
Increasing the attention window from the default 512 to 1024 or 2048 tokens yields modest but consistent improvements in spoken-content metrics, particularly on sWUGGY, sBLiMP, and tSC, while largely preserving acoustic consistency. The largest window achieves the best results on tSC and perplexity, and slightly improves UTMOS and speaker similarity, suggesting that WavSLM benefits from additional context without requiring very large attention windows. Notably, with larger windows WavSLM also achieves lower perplexity than LLaMA-Mimi 1.3B (see \cref{tab:utmos_sim_ppl})

In contrast, increasing the chunk size from the default 4 to 8 or 16 improves generation speed but leads to substantial degradations in both likelihood-based and generation-based metrics. Larger chunks reduce UTMOS and speaker similarity and increase perplexity, indicating a loss in acoustic fidelity and linguistic coherence. This suggests that overly long chunks, well above the chunk size used by the tokenizer, are detrimental to modeling performance.

\section{Conclusion}
We introduced {WavSLM}, a speech language modeling framework that revisits the simplicity of autoregressive text language models in the speech domain. By combining a WavLM-based initialization with a streamable neural speech codec, WavSLM distills rich speech representations into an autoregressive next-chunk predictor operating over a single discrete token stream. This design enables causal and streamable speech language modeling while jointly preserving semantic and acoustic information.
Despite its simplicity and speech-only training, WavSLM achieves competitive performance across semantic, acoustic, and generative benchmarks, pointing toward a promising direction for simpler, more efficient, and more scalable SLMs.

\section{Acknowledgments}
We gratefully acknowledge the support of NSERC, the Digital Research Alliance of Canada (alliancecan.ca), Translated (Imminent Program), and Apple (Seed Grant) through research funding, computing resources, and donations.

\section{Generative AI Use Disclosure}
All authors take full responsibility and accountability for the content of this paper and consent to its submission. 
Generative AI tools were used solely for minor editing and language polishing to improve clarity and fluency of the manuscript.

\bibliographystyle{IEEEtran}
\bibliography{refs}

\end{document}